\documentclass[letterpaper, 10 pt, conference]{ieeeconf}  

\IEEEoverridecommandlockouts                              
\usepackage{amsmath,graphicx,algorithm,algorithmic}

\usepackage{amsmath,amssymb,amsfonts,latexsym,bbm,xspace,graphicx,float,mathtools,paralist,verbatim,xcolor}
\usepackage{url}
\PassOptionsToPackage{hyphens}{url}\usepackage{hyperref}

\newcommand{\thresh}{\gamma}
\newcommand{\prob}{\mathbb{P}}

\newcommand{\probe}{\alpha}
\newcommand{\response}{\beta}
\newcommand{\utility}{u}

\newcommand{\reals}{\mathbb{R}}
\newcommand{\dataset}{\mathcal{D}}

\renewcommand{\time}{k}
\newcommand{\horizon}{K}

\newcommand{\probedim}{m}
\newcommand{\nresponse}{\hat{\response}}
\newcommand{\pertresponse}{\tilde{\response}}
\newcommand{\optpertresponse}{\pertresponse^\ast}
\newcommand{\idresponse}{\response^\ast}

\newcommand{\argmin}{\operatorname{argmin}}
\newcommand{\argmax}{\operatorname{argmax}}
\newcommand{\nonlinb}{g}
\newcommand{\budgetfeas}{\bar{\nonlinb}}
\newcommand{\timetwo}{l}
\newcommand{\margin}{\psi}

\newcommand{\pert}{\delta}
\newcommand{\errprob}{P_{\text{err}}}
\newcommand{\truemargin}{\margin_{g}^{\text{true}}}

\newcommand{\noisecov}{\Sigma}
\newcommand{\margindiff}{\Delta}
\newcommand{\nutility}{\hat{\utility}}
\newcommand{\noptpertresponse}{\hat{\response}^\ast}
\newcommand{\nthresh}{\hat{\thresh}}
\newcommand{\nlambda}{\hat{\lambda}}
\newcommand{\neps}{\hat{\epsilon}}
\newcommand{\eps}{\epsilon}
\newcommand{\SINR}{\operatorname{SINR}}

\newtheorem{definition}{Definition}
\newtheorem{theorem}{Theorem}

\title{Inverse-Inverse Reinforcement Learning.
How to Hide Strategy from an Adversarial Inverse Reinforcement Learner
}
\author{Kunal Pattanayak, Vikram Krishnamurthy and Christopher Berry\thanks{V. Krishnamurthy and K. Pattanayak are with the School	of Electrical and Computer Engineering, Cornell University, Ithaca,	NY, 14853 USA. e-mail: vikramk@cornell.edu, kp487@cornell.edu. C. Berry is with Lockheed Martin Advanced Technology Laboratories, Cherry Hill, NJ, 08002 USA. e-mail: christopher.m.berry@lmco.com. This research was supported in part by a research contract from  Lockheed Martin and  the Army Research Office grant W911NF-21-1-0093.}}

\begin{document}
%
\maketitle
\begin{abstract}
Inverse reinforcement learning (IRL) deals with estimating an agent's utility function from its actions. In this paper, we consider how an agent can hide its strategy and mitigate an adversarial IRL attack;  we call this inverse IRL (I-IRL). {\em How should the decision maker choose its response to ensure a poor reconstruction of its strategy by an adversary performing IRL to estimate the agent's strategy?} This paper comprises four results: First, we present an adversarial IRL algorithm that estimates the agent's strategy while controlling the agent's utility function. Our second result for I-IRL result spoofs the IRL algorithm used by the adversary. Our I-IRL results are based on revealed preference theory in micro-economics. The key idea is for the agent to deliberately choose sub-optimal responses that sufficiently masks its true strategy. Third, we give a sample complexity result for our main I-IRL result when the agent has noisy estimates of the adversary specified utility function. Finally, we illustrate our I-IRL scheme in a radar problem where a meta-cognitive radar is trying to mitigate an adversarial target. 
\end{abstract}
%
%

\section{Introduction}
\begin{figure*}
    \centering
    \includegraphics[width=\linewidth]{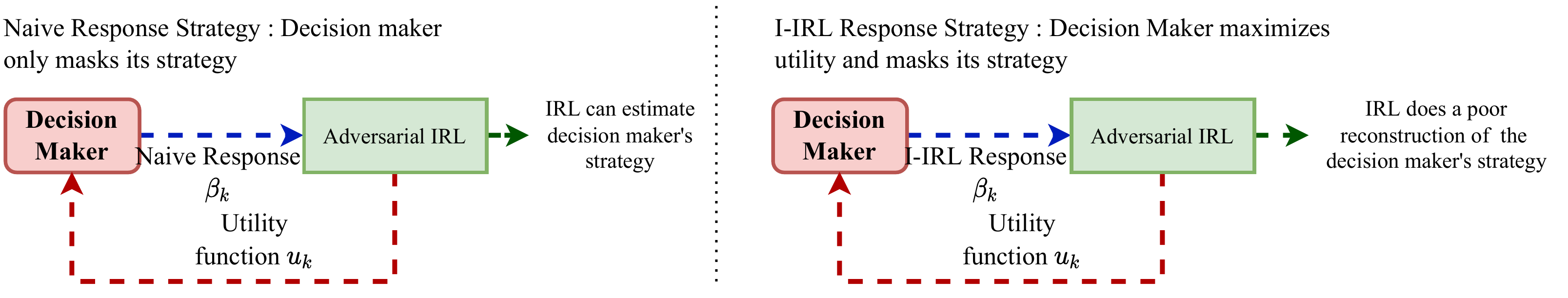}     \caption{Schematic of the I-IRL scheme for masking the strategy of a cognitive decision maker from adversarial IRL. \\
    {\em Naive response strategy (Left):} The adversary sends a sequence of probe signals to the decision maker and records its responses. The probe signal parameterizes the decision maker's utility function. If the decision maker chooses the optimal (naive) response that maximizes its utility function subject to its capability constraint, its capability can be estimated by the adversary using Theorem~\ref{thrm:rp_constraint}.\\
    {\em Adversarial inverse IRL strategy (Right):} If the decision maker is aware that the adversary is trying to estimate its capability, the decision maker deliberately chooses sub-optimal responses via Theorem~\ref{thrm:irp_constraint} to mitigate the adversary. The key idea is to ensure a poor reconstruction of the decision maker's constraint by the adversary by minimally perturbing its naive responses. }
    \label{fig:schematic}
\end{figure*}

This paper studies the interaction between two entities - a smart decision maker and an adversary that aims to estimate the plan of the decision maker; see Fig.\,\ref{fig:schematic} for a schematic representation. The adversary sends adversarial probes to the decision maker and controls the decision maker's utility function. In turn, the decision maker's response maximizes its utility function subject to the decision maker's budget constraint. The adversary's intent is to estimate the budget constraints of the decision maker. If the decision maker knows of the adversarial attack, how should the decision maker tweak its responses to spoof the adversary?

We formulate this interaction between the decision maker and adversary as an
{\em inverse-inverse reinforcement learning} problem.
Reinforcement learning (RL)~\cite{SUT18,KA18} deals with learning the optimal decision making strategy  by observing the response to a control input.
{\em Inverse}  reinforcement learning (IRL)~\cite{ABB01,Afr67,Var83} is the problem of reconstructing the utility function of a decision maker by observing its actions. Inverse IRL (I-IRL) is a natural extension of IRL: {\em If a decision maker knows that an
adversary is using an IRL algorithm to reconstruct its strategy by observing its utility function,  how should the decision maker deliberately tweak its response to mitigate the IRL algorithm?}\footnote{Though not discussed in this paper, an immediate extension is to formulate the decision maker-adversary interaction as a game, and is a topic of current research.}

{\em Outline and Main Results.}
This paper considers a revealed preference based adversarial IRL scheme to estimate the decision maker's strategy. Sec.\,\ref{sec:background} covers the key results from revealed preference theory in micro-economics. Revealed preference studies non-parametric detection of constrained utility maximization behavior. Theorem~\ref{thrm:rp} in Sec.\,\ref{sec:background} presents a feasible test for identifying constrained utility maximization behavior, and generates a set-valued estimate of the decision maker's utility function. Before we address the problem of I-IRL for hiding strategy, we state Theorem~\ref{thrm:rp_constraint}, an IRL algorithm for estimating the strategy (budget constraint) of a decision maker when its utility function is known to the adversary. While Theorem~\ref{thrm:rp} is well known in literature for estimating a utility function, Theorem~\ref{thrm:rp_constraint} is new. Next, in Sec.\,\ref{sec:iirl_constraint}, we state our main result, Theorem~\ref{thrm:irp_constraint}. If the decision maker knows an adversary is using Theorem~\ref{thrm:rp_constraint} to reconstruct, it deliberately chooses sub-optimal responses that minimally violates its strategic constraints using the I-IRL scheme of Theorem~\ref{thrm:irp_constraint} to obfuscate the adversarial attack. Sec.\,\ref{sec:iirl_constraint} also presents a finite sample complexity result, Theorem~\ref{thrm:finitesample} that upper bounds the probability that the I-IRL scheme of Theorem~\ref{thrm:irp_constraint} fails when the decision maker has noisy measurements of the adversary specified utility functions. Finally, Sec.\,\ref{sec:numerical_results} illustrate our I-IRL result for hiding strategy in a radar problem, wherein a cognitive radar is trying to mitigate an adversarial target.

\noindent {\bf Related Work.} Our I-IRL result is based on adversarial obfuscation in machine learning. \cite{advobf} provide a comprehensive list of adversarial attacks and robustness to adversarial attacks in machine learning. Our recent work \cite{PKB21}  presents a cognition-masking scheme for a cognitive radar when the adversary has accurate measurements of the radar's response. This paper generalizes \cite{PKB21} in two major ways:
First, we develop IRL results for estimating the decision maker's strategy followed by I-IRL result for masking strategy. Second, we analyze the performance of our I-IRL result in noisy settings via a finite sample complexity test.

This paper comprises a numerical example involving a cognitive radar trying to mitigate an adversarial target.
A cognitive radar~\cite{Hay06,Hay12,BBSJR15}  uses the perception-action cycle of cognition to sense the environment and learn from it relevant information about the target and the environment.
I-IRL for a cognitive radar can be viewed as a form of meta-cognition. Meta-cognition is a sophisticated form of electronic counter countermeasure (ECCM) to electronic countermeasures (ECM) in electronic warfare. \cite{ECCMsurvey} provides a comprehensive list of ECCM techniques.
\cite{ECCM1,ECCM2} propose waveform adaptation schemes to counter barrage jamming. \cite{stealth1,stealth2,stealth3} exploit frequency diversity for radio stealth in multi-target and moving target tracking.
However, meta-cognitive strategies involving deliberate violation of strategy to confuse the adversary's ECM have not been explored previously.
\section{Background. Revealed Preference for Adversarial IRL}
\label{sec:background}
We start by briefly reviewing the key result in the area of revealed preference in  microeconomics theory. Revealed preference studies non-parametric detection of utility maximization behavior. A utility maximizer is defined as:
\begin{definition}[\cite{Afr87,FM09}]\label{def:UM}
An agent is  a {\em utility maximizer} if for every constraint $\nonlinb_\time(\response)\leq 0$, the  response $\response_\time \in \reals^\probedim_+$ satisfies
\begin{equation}
 \response_\time\in \argmax \utility(\response),~\nonlinb_\time(\response)\leq 0
\label{eqn:utilitymaximization}
\end{equation}
where $\utility(\response)$ is a {\em monotone} utility function.
 \end{definition}

Definition~\ref{def:UM} rationalizes consumer behavior in economics. The constraint $\nonlinb_\time(\response)\leq 0$ in \eqref{eqn:utilitymaximization} is the budget faced by the consumer and  $\response_\time$ is the consumer's consumption vector. In the special case when $\nonlinb_\time(\response)$ is linear, that is, $\nonlinb_\time(\response) = \probe_\time'\response-1$, $\probe_\time$ can be interpreted as the price vector faced by the consumer; then $\probe_\time' \response \leq 1$ is a natural budget constraint
for a consumer with 1 dollar. Given a dataset of budget and consumption vectors, the aim in revealed preference is to determine if the consumer is a utility maximizer (rational) that satisfies \eqref{eqn:utilitymaximization}.  Indeed, the budget constraint $\probe_\time' \response \leq 1$ is without loss of generality, and can be replaced by $\probe_\time' \response \leq c$ for any positive constant $c$.



\subsection{Adversarial IRL for Identifying Utility Function}
The key result in revealed preference is Afriat's theorem ~\cite{Afr67,Var12}. Afriat's theorem assumes a linear budget and specifies a set of linear inequalities that are both necessary and sufficient for a time series of constraints and responses to be consistent with utility maximization behavior~\eqref{eqn:utilitymaximization}. \cite{FM09} propose a utility maximization test that generalizes Afriat's Theorem to non-linear budgets and is the key IRL algorithm used by the adversary in this paper:
\begin{theorem}[Test for utility maximization \cite{FM09}] Given a sequence of constraints and responses $\dataset=\{(\nonlinb_\time(\response)\leq 0,\response_\time)\}_{\time=1}^ \horizon$. Suppose the constraint is active at $\response_\time$, i.e.\,, $\nonlinb_\time(\response_\time)=0~\forall \time$. Then, the following statements are equivalent:
	\begin{compactenum}
	\item There exists a monotone, continuous utility function that satisfies \eqref{eqn:utilitymaximization}.
	\item  There exist positive reals $\{u_t,\lambda_t\}_{t=1}^\horizon$ such that the following inequalities are feasible:
			\begin{equation}
				u_s-u_t-\lambda_t \nonlinb_t(\response_s) \leq 0 \; \forall t,s\in\{1,\dots,\horizon\}.\
				\label{eqn:AfriatFeasibilityTest}
			\end{equation}
			The monotone utility function
			given by
			\begin{equation}
				\utility(\response) = \underset{t\in \{1,2,\dots,\horizon\}}{\operatorname{min}}\{u_t+\lambda_t \nonlinb_t(\response)\}
				\label{eqn:estutility}
            \end{equation}
            constructed using feasible $u_t$ and $\lambda_t$ \eqref{eqn:AfriatFeasibilityTest} rationalizes $\mathcal{D}$.
          \item The data set $\mathcal{D}$ satisfies the Generalized Axiom of Revealed Preference (GARP), namely, for any $k\in\{1,2,\ldots,\horizon\}$, the following implication holds:
          \begin{equation}\label{eqn:GARP}
           \nonlinb_t(\response_{t+1})\leq \nonlinb_t(\response_t) \quad \forall t\leq k-1 \implies \nonlinb_k(\response_{1})\geq \nonlinb_k(\response_{k}).
          \end{equation}
	\end{compactenum}
\label{thrm:rp}
\end{theorem}
Theorem~\ref{thrm:rp} tests for economics-based rationality; its  remarkable property is that it gives a {\em necessary and sufficient condition} for a agent to be a utility maximizer based on the agent's input-output response.
The feasibility of the set of inequalities (\ref{eqn:AfriatFeasibilityTest}) can be checked using a linear programming solver; alternatively GARP can be checked  using Warshall's algorithm with $O(\horizon^3)$ computations~\cite{Var06,Var82}. Theorem~\ref{thrm:rp} can be viewed as set-valued system identification of an \emph{argmax} system; set-valued since (\ref{eqn:estutility}) yields a set of utility functions that rationalize the finite dataset $\dataset$.

{\em Key Idea for I-IRL: Manipulating the Goodness-of-fit of revealed preference test \eqref{eqn:AfriatFeasibilityTest}.} Theorem~\ref{thrm:rp} also constructs a set-valued estimate~\eqref{eqn:estutility} of the utility function $\utility$ using the solution of the set of feasibility inequalities \eqref{eqn:AfriatFeasibilityTest}. The estimated utility function \eqref{eqn:estutility} is ordinal since any positive monotone increasing transformation of~\eqref{eqn:estutility} also satisfies Theorem~\ref{thrm:rp}. We make two observations here that are crucial for our I-IRL results in Sec.\,\ref{sec:iirl_constraint}:\\
1. Since the feasibility of \eqref{eqn:AfriatFeasibilityTest} is necessary for utility maximization, the scalars $\utility(\response_\time),\lambda_\time$ satisfy the revealed preference test of \eqref{eqn:AfriatFeasibilityTest}, where $\lambda_\time$ solves $\lambda_\time \nabla \nonlinb_\time(\response_\time) = \nabla \utility(\response_\time)$. Due to the monotonicity of $u,\nonlinb_\time$ and the assumption that the constraint is active ($\nonlinb_\time(\response_\time)=0~\forall \time$), $\lambda_\time$ is well-defined.\\
2. The reconstructed utility function~\eqref{eqn:estutility} is a point-wise minimum of monotone functions parameterized by positive reals $\{\utility_\time,\lambda_\time\}$ that satisfy \eqref{eqn:AfriatFeasibilityTest}. Hence, one can at best recover a lower envelope of the true utility function $u$ that matches the function value and gradient value at the points $\response_\time,\time=1,2,\ldots,\horizon$ using Theorem~\ref{thrm:rp}. In other words, the closest approximation $\utility_{\text{best}}$ to the decision maker's utility $u$ via the reconstruction procedure of \eqref{eqn:estutility} is given by:
\begin{align}\label{eqn:best_estutility}
    \utility_{\text{best}}(\response) = & \underset{\time\in\{1,2,\ldots,\horizon\}}{\min} \{u(\response_\time) + \lambda_\time \nonlinb_\time(\response)\},\\
    \text{ where }& \lambda_\time\nabla \nonlinb_\time(\response_\time) = \nabla \utility(\response_\time).\nonumber
\end{align}
Also, one can show that $\utility_{\text{best}}$~\eqref{eqn:best_estutility} is the least squares estimate of $\utility$:
\begin{align}
    \{\utility(\response_\time),\lambda_\time\} =  \underset{\bar{\lambda}_\time,\utility_\time\geq 0}{\argmin} \int_{\mathcal{S}} \left( \utility(\response) -  \underset{t}{\operatorname{min}}\{u_t+\bar{\lambda}_t \nonlinb_t(\response)\} \right)^2 d\response, \label{eqn:mlse_util}
\end{align}
for any  compact set $\mathcal{S} \subset \reals_+^\horizon$, where $\lambda_\time$ is defined in \eqref{eqn:best_estutility}.

Our key idea for I-IRL is to perturb the response sequence $\{\response_\time\}$ so that the closest IRL estimate~\eqref{eqn:best_estutility} of the decision maker's system parameters passes the revealed preference test of \eqref{eqn:AfriatFeasibilityTest} by a low margin, where the margin is defined by:
\begin{equation}\label{eqn:margin_rp}
    \margin_{u}(\{\response_\time,\nonlinb_\time\}) = \max_{j,k} u(\response_j)-u(\response_k)-\lambda_k \nonlinb_k(\response_j),
\end{equation}
where $~\lambda_k\nabla \nonlinb_\time(\response_\time) = \nabla \utility(\response_\time)$. The margin \eqref{eqn:margin_rp} is a measure of goodness-of-fit~\cite{Var83} of the revealed preference inequalities \eqref{eqn:AfriatFeasibilityTest}. Hence, a utility function that passes \eqref{eqn:AfriatFeasibilityTest} with a large margin is a high-confidence point utility estimate for the adversary and vice versa.

Below, we present a revealed preference test, Theorem~\ref{thrm:rp_constraint}, that tests for feasible budget constraints estimating the decision maker's budget constraint when its utility function is known. The aim of our key I-IRL result of Theorem~\ref{thrm:irp_constraint} in Sec.\,\ref{sec:iirl_constraint} is to ensure that the closest IRL estimate of the decision maker's constraint sequence $\{\nonlinb_\time(\cdot)\}$ passes the revealed preference test of Theorem~\ref{thrm:rp_constraint} by a low margin~\eqref{eqn:margin_rp}.


\subsection{Adversarial IRL for Identifying Strategy }\label{sec:irl_constraint}
Theorem~\ref{thrm:rp} achieves IRL when an adversarial learner wants to estimate the decision maker's utility function and knows the decision maker's budget constraint sequence (strategy). We now consider the scenario where the adversary's probes parametrize the decision maker's utility, and the adversary's aim is to estimate the unknown budget constraint sequence $\{\nonlinb_\time(\response)\leq 0\}$ (strategy) of the decision maker. Below, we present Theorem~\ref{thrm:rp_constraint}, a revealed preference test for the existence of feasible budget constraints when the utility function and decision maker's response is observed by the adversary.
\begin{theorem}[IRL for Identifying Strategy] Given a time sequence of adversary controlled utility functions and decision maker's responses $\dataset=\{(\utility_\time,\response_\time)\}_{\time=1}^\horizon$. Suppose the decision maker faces a budget constraint of the form $\nonlinb(\response)-\thresh_\time\leq 0$ for every $\time$.
Then, the following statements are equivalent:
	\begin{compactenum}
	\item There exists a sequence of monotone continuous capability constraints $\{\nonlinb_\time(\response)\leq 0\}$ that satisfy \eqref{eqn:utilitymaximization}:
	\begin{equation}\label{eqn:utilitymaximization_gen}
	    \response_\time = \argmax~\utility_\time(\response),~\nonlinb_\time(\response)\leq 0
	\end{equation}
	\item There exist positive reals $\{\budgetfeas_\time,\lambda_\time\}_{\time=1}^\horizon$ such that the following inequalities are feasible:
			\begin{equation}
	        	\budgetfeas_s-\budgetfeas_t-\lambda_t~ (\utility_t(\response_s) - \utility_t(\response_t)) \geq 0,~ \forall t,s.
				\label{eqn:AFT_constraint}
			\end{equation}
			The sequence of monotone constraints $\{\nonlinb(\response)- \budgetfeas_\time\leq 0\}$ rationalizes $\dataset$~\eqref{eqn:utilitymaximization}, where budget $\nonlinb$ is given by:
			\begin{equation}
				\nonlinb(\response) = \underset{t\in \{1,2,\dots,\horizon\}}{\operatorname{max}}\{\budgetfeas_t+\lambda_t~(\utility_t(\response) - \utility_t(\response_t))\}.
				\label{eqn:estbudget}
            \end{equation}
          \item The data set $\{\utility_t(\response_t) - \utility_t(\cdot),\response_t\}$ satisfies GARP~\eqref{eqn:GARP}.
	\end{compactenum}
\label{thrm:rp_constraint}
\end{theorem}
The proof of Theorem~\ref{thrm:rp_constraint} is omitted for brevity; see \cite{PK21unifying} for a more elaborate discussion. At first sight, Theorem~\ref{thrm:rp_constraint} appears to be a dual statement to the optimization in Theorem~\ref{thrm:rp}. Instead of testing for a rationalizing utility given a sequence of known  budget constraints,  Theorem~\ref{thrm:rp_constraint} tests for a rationalizing sequence of budget constraints given the utility function and does not use duality in the proof.

In complete analogy to Theorem~\ref{thrm:rp}, the feasibility inequality of \eqref{eqn:AFT_constraint} is necessary and sufficient for the existence of a sequence of constraints that rationalizes the sequence of utility functions and responses. In complete analogy to \eqref{eqn:best_estutility}, we now define $\nonlinb_{\text{best}}$, the closest approximation (upper envelope) to the true budget $\nonlinb$ reconstructed via \eqref{eqn:AFT_constraint}:
\begin{align}
    \nonlinb_{\text{best}}(\response) = & \underset{\time\in\{1,2,\ldots,\horizon\}}{\max} \{\thresh_\time + \lambda_\time (\utility_\time(\response) - \utility_\time(\response_\time)) \},
\end{align}
 where $\lambda_\time\nabla \nonlinb_\time(\response_\time) = \nabla \utility(\response_\time)$.
Analogous to \eqref{eqn:margin_rp}, we define the margin with which the true budget $\nonlinb$ passes the revealed preference test~\eqref{eqn:AFT_constraint} of Theorem~\ref{thrm:rp_constraint}:
\begin{align}
    \margin_g(\{\response_\time,\utility_\time,\thresh_\time\}) = & \underset{j,k}{\min} ~ \nonlinb(\response_j)-\nonlinb(\response_k)-\lambda_k~ (\utility_k(\response_j) - \utility_k(\response_k)),\nonumber\\
    \text{where } & \lambda_k \nabla \utility_k(\response_k) = \nabla \nonlinb(\response_k). \label{eqn:margin_rp_con}
\end{align}
In our I-IRL results in the next section, our key objective will be to minimally perturb the response sequence $\{\response_\time\}$ so that $\margin_g(\cdot)$ lies below a pre-specified threshold.

Theorem~\ref{thrm:rp_constraint} assumes the elements in the sequence of constraints $\{\nonlinb(\response)-\thresh_\time\}$ differ only by a scalar shift. This assumption can indeed be relaxed to allow any sequence of budget constraints. But the {\em reconstructed} constraints~\eqref{eqn:estbudget} are restricted 
to the space of monotone piece-wise linear convex functions identical up to a constant. Hence, any constraint that lies outside this space is non-identifiable.


\section{Inverse IRL (I-IRL) for Masking Decision Maker's Strategy}
\label{sec:iirl_constraint}
Sec.\,\ref{sec:background} presents IRL algorithms that an adversary uses to estimate the decision maker's strategy. If the decision maker is aware of the adversarial attack, how should it choose its responses to mask the strategy from the adversary? Below, we present our main I-IRL result, Theorem~\ref{thrm:irp_constraint}. In Sec.\,\ref{sec:finitesample}, we give a finite sample result for Theorem~\ref{thrm:irp_constraint} that upper bounds the probability the I-IRL scheme of Theorem~\ref{thrm:irp_constraint} fails when the decision maker's utility function is corrupted by additive noise.


\subsection{Main Result. I-IRL for Adversarial IRL (Theorem~\ref{thrm:rp_constraint})}
\begin{theorem}[I-IRL for Masking Strategy] Suppose response $\idresponse_\time$ maximizes the adversary controlled utility function $\utility_\time$ subject to budget constraint $\nonlinb(\response)\leq \thresh_\time$ for time $\time=1,2,\ldots,\horizon$. Also, suppose the adversary uses Theorem~\ref{thrm:rp_constraint} to reconstruct the decision maker's budget $\nonlinb(\cdot)$. Then, the I-IRL response sequence $\{\optpertresponse_\time\}$ of the decision maker for masking its budget $\nonlinb(\cdot)$ is given by:
\begin{equation}\label{eqn:iirl_response}
 \pertresponse_{\time}^\ast = \argmax_{\response}u_\time(\response),~\nonlinb_\time(\response)\leq \gamma_\time^\ast,
 \end{equation}
 where the violated budget thresholds $\{\thresh_\time^\ast\}$ solve the following optimization problem:
 \begin{align}\{\gamma_\time^\ast\}&=\underset{\tilde{\thresh}_{1:\horizon}}{\argmin} \sum_{\time=1}^\horizon  \|\tilde{\gamma}_\time - \gamma_\time\|_2^2, \label{eqn:irp_budget}\\ \margin_{\nonlinb}(\{\tilde{\response}_\time,\utility_\time,&\tilde{\thresh}_\time\}) \leq (1-\eta)~\margin_{\nonlinb}(\{\idresponse_\time,\utility_\time,\thresh_\time\}),\label{eqn:constraint_lowmargin_budget}\\
 \tilde{\response}_\time & =  \argmax_{\response} \utility_\time(\response),~\nonlinb(\response)\leq \tilde{\thresh}_\time. \nonumber
\end{align}
In \eqref{eqn:constraint_lowmargin_budget}, $\eta\in[0,1]$ is a pre-defined scalar that parameterizes the extent of strategy masking for I-IRL.
\label{thrm:irp_constraint}
\end{theorem}

Theorem~\ref{thrm:irp_constraint} is the main I-IRL result of this paper. The decision maker's I-IRL response maximizes $\utility_\time(\cdot)$ subject to a violated budget constraint $\nonlinb(\response)\leq \thresh_\time^\ast$. The I-IRL scheme of Theorem~\ref{thrm:irp_constraint} optimally trades-off between minimizing performance loss of the decision maker due to constraint violation, and spoofing the IRL algorithm of the adversary (by decreasing the margin of the IRL feasibility test \eqref{eqn:AFT_constraint}). \vspace{0.1cm}\\

\noindent {\em Discussion.} Recall from Sec.\,\ref{sec:irl_constraint} that $\margin_{\nonlinb}(\{\idresponse_\time,\utility_\time,\thresh_\time\})$ \eqref{eqn:constraint_lowmargin_budget} is the margin with which the budget $\nonlinb(\response)$ and the naive response sequence $\{\idresponse_\time\}$ pass the revealed preference test of \eqref{eqn:AFT_constraint}. Due to the necessity of \eqref{eqn:AFT_constraint} for utility maximization behavior, the RHS in \eqref{eqn:constraint_lowmargin_budget} is non-negative and $0$ iff $\eta=0$. The LHS term in \eqref{eqn:constraint_lowmargin_budget} is the margin with which the I-IRL response sequence $\{\optpertresponse_\time\}$ and violated budget constraints pass \eqref{eqn:AFT_constraint}. Hence, \eqref{eqn:irp_budget} computes the minimum violation that reduces the margin with the I-IRL response pass the revealed preference test of \eqref{eqn:AFT_constraint} by a factor of $1/(1-\eta)$.

It is straightforward to show the minimum violation of constraints \eqref{eqn:irp_budget} is monotone in the parameter $\eta$. If $\eta=0$, the I-IRL response $\{\optpertresponse_\time\}$ is identical to the naive response $\{\idresponse_\time\}$ and the minimum violation of budget is $0$. On the other extreme, setting $\eta=1$ requires maximal violation of the budget constraints $\{\nonlinb(\response)\leq \thresh_\time\}$ since $\margin_{\nonlinb}(\{\tilde{\response}_\time,\utility_\time,\tilde{\thresh}_\time\})\leq 0$ \eqref{eqn:constraint_lowmargin_budget}   implies the I-IRL response and decision maker's budget fail the revealed preference test of Theorem~\ref{thrm:rp_constraint}. We illustrate the I-IRL result in the next section via a radar example; see Fig.\,\ref{fig:radar_example} for the simulation result. We show how the radar can mitigate an adversarial target by deliberately choosing sub-optimal waveform that mitigates its signal-to-interference-plus-noise ratio (SINR), while ensuring a poor reconstruction of its budget by the adversarial target.

\subsection{Finite Sample Complexity for I-IRL (Theorem~\ref{thrm:irp_constraint})}\label{sec:finitesample}
In the previous sections, we assumed both the adversary and the decision maker had accurate measurements of the response and the utility functions. In this section, we assume the decision maker's measurements of the utility function is noisy, and the noise is modeled as a random linear perturbation. The key question we address is:\\
{\em Given a finite sequence of I-IRL responses to noisy utility functions $\utility_\time(\response)+\pert_\time'\response$, what is probability that the decision maker effectively masks its strategy from the adversary?}\vspace{0.1cm}\\
Let us now formalize the above question. Let $\margin_g^{\text{true}}=\margin_{\nonlinb}(\{\idresponse_\time,\utility_\time,\thresh_\time\})$~\eqref{eqn:margin_rp_con} denote the margin with which the naive response sequence $\{\idresponse_\time\}$~\eqref{eqn:utilitymaximization} passes the revealed preference test of Theorem~\ref{thrm:rp_constraint}. We want to bound the following error probability for I-IRL in Theorem~\ref{thrm:irp_constraint}:
\begin{equation}\label{eqn:finite_err_prob}
    \errprob=\prob_{\pert_{1:\horizon}}\left(\margin_{\nonlinb}(\{\optpertresponse_\time,\utility_\time(\cdot) + \pert_\time'(\cdot),\thresh_\time^\ast\}) \geq (1-\eta)~\margin_g^{\text{true}}\right)
\end{equation}
Recall from Theorem~\ref{thrm:irp_constraint} that our I-IRL aim is to ensure the margin of the revealed preference test~\eqref{eqn:AFT_constraint} lies under a threshold. In \eqref{eqn:finite_err_prob}, $\errprob$ is the probability with which the constraint~\eqref{eqn:irp_budget} in Theorem~\ref{thrm:irp_constraint} fails. In simple terms, $\errprob$ is the probability of the event that the margin with which the I-IRL response satisfies the inequalities~\eqref{eqn:AFT_constraint} in Theorem~\ref{thrm:rp_constraint} exceeds the margin threshold $(1-\eta)\truemargin$.

We assume the following for Theorem~\ref{thrm:finitesample}:
\begin{compactitem}
\item[(A1)] The adversary controlled utility function $\utility_\time$ is monotone, concave and Lipschitz continuous with Lipschitz constant $L$.
\item[(A2)] The decision maker has a noisy estimate $\nutility_\time=\utility_\time(\response) + \pert_\time(\response)$ of the adversary controlled utility function $\utility_\time(\response)$. The linear perturbation vector $\pert_\time$ is a Gaussian zero mean random vector with covariance $\noisecov$.
\item[(A3)] Let $\Delta(\nonlinb,\{\response_\time,\utility_\time,\thresh_\time\})$ denote the range with which $\nonlinb,\{\response_\time,\utility_\time,\thresh_\time\}$ pass the revealed preference test of \eqref{eqn:AFT_constraint}:
\begin{align}
&\margindiff(\nonlinb,\{\response_\time,\utility_\time,\thresh_\time\})  =   \max_{j,k} \eps_{j,k}  - \min_{j,k} \eps_{j,k},~\text{where}\nonumber\\
&\eps_{j,k} =  \thresh_j-\thresh_k-\lambda_k~ (\utility_k(\response_j) - \utility_k(\response_k)),\nonumber\\
&  \lambda_\time\nabla\utility_\time(\response_\time) = \nabla \nonlinb(\response_\time).\nonumber \label{eqn:spread_rp_con}
\end{align}
The random variable $\Delta(\nonlinb,\{\nresponse_\time,\nutility_\time,\nthresh_\time\})\leq \Delta_{\max}$ a.s.\,, where $\nresponse_\time$ and $\nthresh_\time$ are the decision maker's I-IRL response~\eqref{eqn:iirl_response} and constraint threshold~\eqref{eqn:irp_budget} due to noisy utility function $\nutility_\time$ measured by the decision maker.
\item[(A4)] The random variable $\frac{\max_k \{||\nabla \utility_k(\nresponse_k) ||_2^2/||\nabla \nonlinb(\nresponse_\time)||_2\}}{\min_{j,k}\{||\nabla \utility_k(\nresponse_k) - \nabla \utility_k(\nresponse_j)||_2^2\}}$ is upper bounded almost surely by $\kappa>0$.
\end{compactitem}

We are now ready our finite sample complexity result for I-IRL (Theorem~\ref{thrm:irp_constraint}).
\begin{theorem}[Finite Sample Complexity for I-IRL]\label{thrm:finitesample} Consider the decision maker choosing I-IRL responses according to \eqref{eqn:iirl_response} in Theorem~\ref{thrm:irp_constraint} in response to noisy utility functions controlled by the adversary. Let $\idresponse_\time$ denote the naive response of the decision maker at time $\time$ that maximizes the noise-less utility $\utility_\time$ subject to budget constraint $\nonlinb(\response)\leq\thresh_\time$. Suppose assumptions (A1)-(A4) hold. Then:
\begin{equation}\label{eqn:upperboundfinite}
    \errprob \leq \boldsymbol{\phi}^\horizon\left(\frac{2L\Delta_{\max}\kappa}{\sqrt{\operatorname{Tr}(\Sigma)}}\right)
\end{equation}
where $\errprob$ is the error probability for I-IRL (Theorem~\ref{thrm:irp_constraint}) defined in \eqref{eqn:finite_err_prob} and  $\boldsymbol{\phi}(\cdot)$ is the cdf of the standard normal distribution.
\end{theorem}
The proof of Theorem~\ref{thrm:finitesample} is in the appendix.

\section{Example. I-IRL for Meta-cognitive radar}\label{sec:numerical_results}

Theorem~\ref{thrm:irp_constraint} specified the procedure for a decision maker to effectively mask its cognition from an adversary.
Here, we apply our I-IRL result to the problem of a cognitive radar optimizing waveform based on the SINR of the adversarial target measurement \cite{KPGKR21}.
The adversary observes the radar over $\time=1,2,\ldots,\horizon$ time epochs.
At the $\time^{th}$ epoch, the adversary probe the radar with an interference vector $\alpha_\time\in\reals^M$.
The radar responds with waveform $\beta_\time\in\reals_{+}^M$, which maximizes its SINR while satisfying a linear budget constraint $p'\response\leq p_k$.
The SINR of the radar given probe $\probe$ and response $\response$ is defined as
\begin{equation}
\label{eqn:SINR_def}
\SINR(\alpha,\beta) = \frac{\beta^{'}Q\beta}{\beta^{'}P(\alpha)\beta + \zeta}.
\end{equation}
In~(\ref{eqn:SINR_def}), the radar's signal power (numerator) and interference power (first term in denominator) are assumed to be quadratic forms of $Q,P(\probe)$ respectively, where $Q,P(\alpha)\in\reals^{M\times M}$ are positive definite matrices known to the adversary.
The term $\zeta>0$ is the noise power.

\begin{figure}
    \centering
    \includegraphics[width = 0.8\columnwidth]{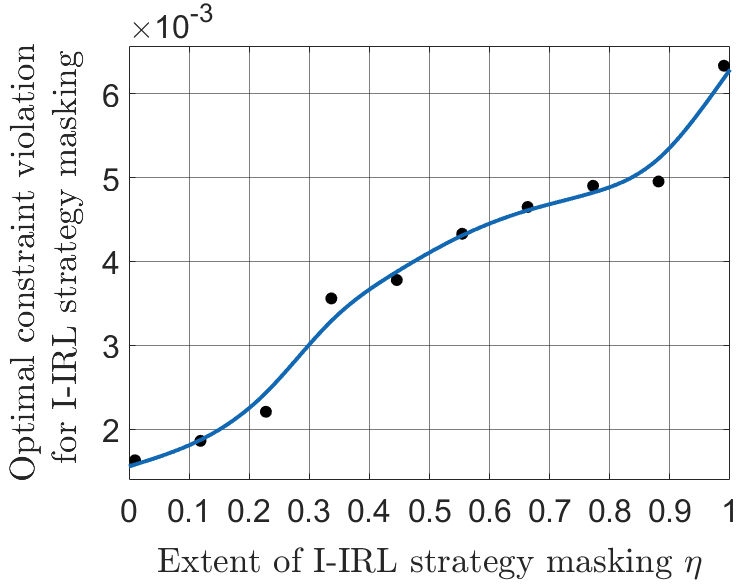}
    \caption{I-IRL for masking the strategy of a cognitive radar: Small deliberate constraint violation of the radar (vertical axis) results in large performance loss (extent of strategy masking $\eta$) of the adversarial IRL algorithm (horizonal axis). $\eta=0$ corresponds to zero strategy masking, and $\eta=1$ corresponds to complete strategy masking by the decision maker. As expected, the optimal deliberate constraint violation by the decision maker increases with $\eta$.
    }
    \label{fig:radar_example}
\end{figure}

Having defined the SINR above in \eqref{eqn:SINR_def}, we now formalize the radar's naive response $\response_\time$ given probe $\probe_\time$, $\time=1,2,\ldots$ as the solution of the following optimization problem.
\begin{align}
    \beta_\time &\in\argmax_{\beta} \operatorname{SINR}(\alpha_\time,\beta)\nonumber\\
    \text{s.t. }&p'\response\leq p_\time,\label{eqn:radar_opt_nonlinear}
\end{align}
In \eqref{eqn:radar_opt_nonlinear}, $p(i)\response(i)$ is the cost of transmitting signal power $\response(i)$ on the $i^\text{th}$ waveform.
Clearly, the above setup falls under the non-linear utility maximization setup in Definition~\ref{def:UM}. For appropriately chosen matrices (see \cite{KPGKR21}), the utility in \eqref{eqn:radar_opt_nonlinear} can be shown to be monotonically increasing in $\response$.
In summary, we have justified how the radar should employ Theorem~\ref{thrm:irp_constraint} to deliberately choose sub-optimal waveforms so that the radar's budget $g$ reduces the margin that the I-IRL response passes the revealed preference test in \ref{eqn:AFT_constraint}.
We illustrate our I-IRL result via a simple numerical example in Fig.\,\ref{fig:radar_example}. For the example, we chose:
\begin{compactitem}
\item Time horizon $\horizon=100$
\item Response dimension $m=6$
\item Budget vector $p=[p(1)~p(2)~\ldots p(m)]$, $p(i)\sim\operatorname{Unif}(1,4)$
\item Extent of strategy masking $\eta$ was varied from $0.05$ to $0.95$ with step size $0.05$
item Matrix $Q=[Q_{i,j}]$, where $Q_{i,i}=5$, $Q_{i,j} = 0$ if $j\neq i$, and $P(\alpha_\time) = [P_{i,j}]$, where $P_{i,i}\sim\operatorname{Unif}(1,3)$ and $P_{i,j} = -0.05$ if $j\neq i$.
\item Noise power $\zeta=1$.
\end{compactitem}
The key observation is that the minimum violation of the radar's strategy increases with increasing extent of budget constraint masking $\eta$.

\section{Conclusion and Extensions}
This paper focuses on masking a decision maker's strategy when probed by an adversarial inverse reinforcement learner. We term this problem inverse-inverse reinforcement learning (I-IRL). If the decision maker knows an adversary is trying to reconstruct its strategy, how should it tweak its responses to hide its strategy? Our main I-IRL result is Theorem~\ref{thrm:irp_constraint}. The key idea is for the decision maker to deliberately choose sub-optimal responses that violates its strategic resource constraints while ensuring the adversary does a poor reconstruction of the decision maker's strategy. Our finite sample result, Theorem~\ref{thrm:finitesample}, upper bounds the probability that our I-IRL result is ineffective in noisy settings; when the decision maker has noisy estimates of the adversary controlled utility functions.


Finally,
a useful  extension of this paper would be to study more general game-theoretic settings where even the adversary knows the radar is trying to mask its cognition. Also, from a counter-counter adversarial perspective, how to design purposeful utility functions that maximize the probability with which the I-IRL result fails.
\section{Appendix}

\subsection{Proof of Theorem~\ref{thrm:rp_constraint}}
{\bf Statement (1)$\implies$(2).}
Fix indices $\timetwo,\time$. Suppose there exist indices $i_1,i_2,\ldots,i_L$ such that
$\utility_\time(\response_\time)-\utility_\time(\response_{i_1})\leq 0,\utility_{i_1}(\response_{i_1})-\utility_{i_1}(\response_{i_2})\leq 0,\ldots,\utility_{i_L}(\response_{i_L})-\utility_{i_L}(\response_\timetwo)\leq 0$. If \eqref{eqn:utilitymaximization} holds, then we must have $\nonlinb(\response_{i_1})\geq\nonlinb(\response_{\time}),\nonlinb(\response_{i_2})\geq\nonlinb(\response_{i_1}),\ldots,\nonlinb(\response_{\timetwo})\geq\nonlinb(\response_{i_L})$, which implies $\nonlinb(\response_\time)\leq \nonlinb(\response_\timetwo)$. Now, assume $\utility_\timetwo(\response_\timetwo)-\utility_\timetwo(\response_\time)<0$. By local non-satiatedness of $\nonlinb$ and continuity of $\utility_\timetwo$, there exists a consumption bundle $\response$ such that
$\utility_\timetwo(\response)>\utility_\timetwo(\response_\timetwo),~\nonlinb(\response)<\nonlinb(\response_\time)\leq\nonlinb(\response_\timetwo)\implies \nonlinb(\response)<\nonlinb(\response_\timetwo)$,
which contradicts our assumption.
and the sequence $\{\response_\time,\utility_\time(\response_\time)-\utility_\time(\response)\},~\time=1,2,\ldots,\horizon$ satisfies GARP~\eqref{eqn:GARP}.

{\bf Statement (2)$\implies$(3).} From the proof of \cite[Proposition 3]{FM09} (see also \cite[Sections 2 and 3]{FST04}), if the sequence $\{\response_\time,\utility_\time(\response_\time)-\utility_\time(\response)\}$ satisfies GARP, then there exist positive scalars $\bar{\nonlinb}_\time,\lambda_\time$ that satisfy the following inequality.
\begin{equation}\label{eqn:proof_th_2}
    \bar{\nonlinb}_\timetwo- \bar{\nonlinb}_\time + \lambda_\time(\utility_\time(\response_\time)-\utility_\time(\response_\timetwo))\leq 0~\quad \forall~\time,\timetwo.
\end{equation}
Define $\hat{g}_\time = M-\bar{g}_\time$, where $M$ is an arbitrary positive scalar that upper bounds $\bar{\nonlinb}_\time$ for all $\time$. By construction, $\hat{g}_\time>0$. Eq.\,\ref{eqn:proof_th_2} can be further simplified in terms of the variable $\bar{\nonlinb}_\time$ as follows.
\begin{align*}
    & \bar{\nonlinb}_\timetwo- \bar{\nonlinb}_\time + \lambda_\time(\utility_\time(\response_\time)-\utility_\time(\response_\timetwo))\leq 0\\
    \implies & - \bar{\nonlinb}_\timetwo- (-\bar{\nonlinb}_\time) - \lambda_\time(\utility_\time(\response_\time)-\utility_\time(\response_\timetwo))\geq 0\\
    \implies & (M-\bar{\nonlinb}_\timetwo)- (M-\bar{\nonlinb}_\time) + \lambda_\time(\utility_\time(\response_\timetwo)-\utility_\time(\response_\time))\geq 0\\
    \implies & \boxed{\hat{\nonlinb}_\timetwo- \hat{\nonlinb}_\time + \lambda_\time(\utility_\time(\response_\timetwo)-\utility_\time(\response_\time))\geq 0 \equiv \eqref{eqn:AFT_constraint}}.
\end{align*}

Consider the reconstructed cost $\nonlinb(\response)=\max_{\time} \{\bar{\nonlinb}_\time + \lambda_{\time}(\utility_\time(\response)-\utility_\time(\response_\time))\}$. The cost $\nonlinb$ is monotone and continuous since it is a point-wise maximum of monotone continuous functions. Using the fact that inequality \eqref{eqn:AFT_constraint} holds, we have $\nonlinb(\response_\time)=\hat{\nonlinb}_\time$. Hence, the decision maker's budget constraints are given by $\{\nonlinb(\cdot)-\hat{\nonlinb}_\time\leq 0\}$. To see that the above budget constraint sequence rationalizes the sequence $\{\utility_\time,\response_\time\}$, fix index $\time$ and consider consumption bundle $\response$ such that $\nonlinb(\response)\leq \hat{\nonlinb}_\time$. By definition, $0\geq \nonlinb(\response)-\hat{\nonlinb}_\time\geq\lambda_\time(\utility_\time(\response)-\utility_\time(\response_\time)),$ which implies $\utility_\time(\response)\leq \utility_\time(\response_\time)$ since $\lambda_\time>0$. Hence, the budget sequence $\{\nonlinb(\cdot)-\hat{\nonlinb}_\time\leq 0\}$ rationalizes the data.

{\bf Statement (3)$\implies$(1).}
The utility function  $\utility_\time$ is assumed to be locally non-satiated for all $\time$. Hence, every function in the set  $\{\bar{\nonlinb}_\time + \lambda_{\time}(\utility_\time(\response)-\utility_\time(\response_\time)),\time=1,2,\ldots,\horizon\}$ is locally non-satiated. Since $\nonlinb(\response)=\max_{\time} \{\bar{\nonlinb}_\time + \lambda_{\time}(\utility_\time(\response)-\utility_\time(\response_\time))\}$ \eqref{eqn:estbudget} is a point-wise maximum of finitely many locally non-satiated functions, $\nonlinb(\cdot)$ is monotone, continuous and locally non-satiated by construction. $\quad\quad\quad\quad\blacksquare$

\subsection{\em Proof of Theorem~\ref{thrm:finitesample}}
We start by computing the margin with which the I-IRL response of the decision maker passes the feasibility inequalities \eqref{eqn:AFT_constraint} of Theorem~\ref{thrm:irp_constraint}. Let $\nutility_\time(\response)=\utility_\time(\response)+\pert_\time'\response$ denote the noisy utility function estimate available to the decision maker. Let $\{\nresponse_\time\}$ and $\{\nthresh_\time\}$ denote the I-IRL responses and perturbed constraint thresholds computed via \eqref{eqn:iirl_response} and  \eqref{eqn:irp_budget}, respectively, in response to noisy utility functions $\{\nutility_\time\}$. The margin $\margin_{\nonlinb}(\nresponse_\time,\utility_\time,\nthresh_\time)$ is defined as:
\begin{equation}\label{eqn:margin_mismatch}
    \margin_{\nonlinb}(\noptpertresponse_\time,\utility_\time,\nthresh_\time) = \min_{j,k} \underbrace{\nthresh_j - \nthresh_k - \lambda_k (\utility_k(\nresponse_j) - \utility_k(\nresponse_j))}_{=\epsilon_{j,k}},
\end{equation}
where $\lambda_k \nabla\utility_\time(\nresponse_\time) = \nabla \nonlinb(\nresponse_\time)$. If $\nutility_\time$ were the true utility function at time $\time$ generated by the adversary, the margin definition in \eqref{eqn:margin_mismatch} changes to:
\begin{equation}\label{eqn:margin_nomismatch}
    \margin_{\nonlinb}(\nresponse_\time,\nutility_\time,\nthresh_\time) = \min_{j,k} \underbrace{\nthresh_j - \nthresh_k - \nlambda_k (\nutility_k(\nresponse_j) - \nutility_k(\nresponse_j))}_{=\neps_{j,k}},
\end{equation}
where $\nlambda_\time\nutility_\time(\nresponse_\time) = \nabla \nonlinb(\nresponse_\time)$. Observe that by definition~\eqref{eqn:irp_budget}, $\margin_{\nonlinb}(\nresponse_\time,\nutility_\time,\nthresh_\time)= (1-\eta)\truemargin$.
Also, we observe that the margin definitions in \eqref{eqn:margin_mismatch} and \eqref{eqn:margin_nomismatch} differ only in the term involving the utility functions. Our aim is to find necessary conditions for which the event $\{\margin_{\nonlinb}(\nresponse_\time,\utility_\time,\nthresh_\time) \geq (1-\eta)\truemargin\}$ holds, or equivalently, the event $\{\margin_{\nonlinb}(\nresponse_\time,\nutility_\time,\nthresh_\time)\leq \margin_{\nonlinb}(\nresponse_\time,\utility_\time,\nthresh_\time)\}$ holds.

Due to assumption (A3), a necessary condition for the event $\{\margin_{\nonlinb}(\nresponse_\time,\utility_\time,\nthresh_\time) \geq (1-\eta)\truemargin\}$ to hold is $\{\eps_{j,k}\geq \neps_{j,k}-\Delta_{\max},~\forall j,k\}$. Fix indices $j,k$. We wish to bound the term $(\neps_{j,k}-\eps_{j,k})$:
\begin{align}
    & \neps_{j,k} - \eps_{j,k}\nonumber\\
    = & \lambda_k (\utility_k(\nresponse_j) - \utility_k(\nresponse_j)) -  \nlambda_k (\nutility_k(\nresponse_j) - \nutility_k(\nresponse_j))\nonumber\\
    = & \lambda_k (\utility_k(\nresponse_j) - \utility_k(\nresponse_j)) - (\lambda_k + (\nlambda_k-\lambda_k)) (\utility_k(\nresponse_j)\nonumber\\
    & \quad\quad\quad\quad\quad\quad\quad - \utility_k(\nresponse_j) + \pert_k'(\nresponse_j-\nresponse_k))\nonumber\\
    = & - (\nlambda_k-\lambda_k) (\utility_k(\nresponse_j) - \utility_k(\nresponse_j)) -  \lambda_k \pert_k'(\nresponse_j-\nresponse_k)\nonumber\\
    & \quad\quad - (\nlambda_k-\lambda_k) \pert_k'(\nresponse_j-\nresponse_k)\nonumber\\
    = & -(\nlambda_k-\lambda_k)(\utility_k(\nresponse_j) - \utility_k(\nresponse_k) - \nabla \utility_k(\response_k)'(\nresponse_j-\nresponse_k))\nonumber\\
    & \quad \quad (\text{since } \lambda_k \nabla \utility_k(\nresponse_k) = \nlambda_k \nabla \nutility_k(\nresponse_k) )\nonumber\\
    = & (\nlambda_k-\lambda_k)(\underbrace{\utility_k(\nresponse_k) + \nabla \utility_k(\response_k)'(\nresponse_j-\nresponse_k) -\utility_k(\nresponse_j)}_{\geq 0}) \label{eqn:pr1}
\end{align}

\noindent Since $\utility_\time$ is $L$-Lipschitz continuous, we have:
\begin{equation} \label{eqn:pr2}
     \utility_k(\nresponse_k) + \nabla \utility_k(\response_k)'(\nresponse_j-\nresponse_k) - \utility_k(\nresponse_j) \geq \frac{1}{2L} ||\nabla \utility_k(\nresponse_k) - \nabla \utility_k(\nresponse_j)||_2^2
\end{equation}
Since $\lambda_k \nabla \utility_k(\nresponse_k) = \nlambda_k \nabla \nutility_k(\nresponse_k)$ from \eqref{eqn:margin_mismatch}, \eqref{eqn:margin_nomismatch}, we have:
\begin{equation} \label{eqn:pr3}
    \nlambda_k - \lambda_k  = \lambda_k \frac{\pert_k'\nabla \utility_k(\nresponse_k)}{||\nabla \utility_k(\nresponse_k) ||_2^2} = \frac{\pert_k'\nabla \nonlinb(\nresponse_\time)}{||\nabla \utility_k(\nresponse_k) ||_2^2}
\end{equation}
Combining \eqref{eqn:pr1}, \eqref{eqn:pr2} and \eqref{eqn:pr3}, we have:
\begin{align*}
    \neps_{j,k} - \eps_{j,k} & \leq \Delta_{\max}\\
    \implies  \pert_\time'\nabla \nonlinb(\nresponse_\time)& \leq \frac{2L\Delta_{\max}||\nabla \utility_k(\nresponse_k) ||_2^2}{\min_{j,k}\{||\nabla \utility_k(\nresponse_k) - \nabla \utility_k(\nresponse_j)||_2^2\}}
\end{align*}
$\{\pert_k'\nabla \nonlinb(\nresponse_\time)\}$ is a sequence of independent zero mean Gaussian random variables with variance $\{\operatorname{Tr}(\Sigma)||\nabla \nonlinb(\nresponse_\time)||_2^2\}$
Also, notice how the LHS in the above inequality does not depend on the index $j$. Thus, our error probability $\errprob$ can be expressed as:
\begin{align*}
   \errprob = & \prob(\neps_{j,k} - \eps_{j,k} \leq \Delta_{\max},~\forall j,k)\\
    \leq & \prob\left(\pert_\time'\nabla \nonlinb(\nresponse_\time)\leq \pi_k\right)\\
    = & \prod_{k=1}^\horizon \prob\left(\pert_\time'\nabla \nonlinb(\nresponse_\time)\leq \pi_k\right)
\end{align*}
\begin{align*}
    = & \prod_{k=1}^\horizon\boldsymbol{\phi}\left( \frac{2L\Delta_{\max}||\nabla \utility_k(\nresponse_k) ||_2^2/||\nabla \nonlinb(\nresponse_\time)||_2}{\sqrt{\operatorname{Tr}(\Sigma)}\min_{j,k}\{||\nabla \utility_k(\nresponse_k) - \nabla \utility_k(\nresponse_j)||_2^2\}} \right)\\
    \leq & \boldsymbol{\phi}^\horizon\left( \frac{2L \Delta_{\max}\max_k \{||\nabla \utility_k(\nresponse_k) ||_2^2/||\nabla \nonlinb(\nresponse_\time)||_2\}}{\sqrt{\operatorname{Tr}(\Sigma)}\min_{j,k}\{||\nabla \utility_k(\nresponse_k) - \nabla \utility_k(\nresponse_j)||_2^2\}} \right)\\
    = & \boldsymbol{\phi}^\horizon\left(\frac{2L\Delta_{\max}\kappa}{\sqrt{\operatorname{Tr}(\Sigma)}}\right)~(\text{from (A4)}) \quad\quad\quad\quad\blacksquare
\end{align*}
\bibliographystyle{unsrt_abbrv_custom}
\bibliography{refs}

\begin{thebibliography}{10}

\bibitem{SUT18}
R.~S. Sutton and A.~G. Barto.
\newblock {\em Reinforcement learning: An introduction}.
\newblock MIT press, 2018.

\bibitem{KA18}
L.~Kang, J.~Bo, L.~Hongwei, and L.~Siyuan.
\newblock Reinforcement learning based anti-jamming frequency hopping
  strategies design for cognitive radar.
\newblock In {\em 2018 IEEE International Conference on Signal Processing,
  Communications and Computing (ICSPCC)}, pages 1--5. IEEE, 2018.

\bibitem{ABB01}
J.~Abounadi, D.~P. Bertsekas, and V.~Borkar.
\newblock Learning algorithms for {M}arkov decision processes with average
  cost.
\newblock {\em SIAM Journal on Control and Optimization}, 40(3):681--698, 2001.

\bibitem{Afr67}
S.~Afriat.
\newblock The construction of utility functions from expenditure data.
\newblock {\em International economic review}, 8(1):67--77, 1967.

\bibitem{Var83}
H.~Varian.
\newblock Non-parametric tests of consumer behaviour.
\newblock {\em The Review of Economic Studies}, 50(1):99--110, 1983.

\bibitem{advobf}
S.~H. Silva and P.~Najafirad.
\newblock Opportunities and challenges in deep learning adversarial robustness:
  A survey.
\newblock {\em arXiv preprint arXiv:2007.00753}, 2020.

\bibitem{PKB21}
K.~Pattanayak, V.~Krishnamurthy, and C.~Berry.
\newblock How can a cognitive radar mask its cognition?
\newblock {\em arXiv preprint arXiv:2110.08608}, 2021.

\bibitem{Hay06}
S.~Haykin.
\newblock Cognitive radar.
\newblock {\em IEEE Signal Processing Magazine}, pages 30--40, Jan. 2006.

\bibitem{Hay12}
S.~Haykin.
\newblock Cognitive dynamic systems: Radar, control, and radio [point of view].
\newblock {\em Proceedings of the IEEE}, 100(7):2095--2103, 2012.

\bibitem{BBSJR15}
K.~Bell, C.~Baker, G.~Smith, J.~Johnson, and M.~Rangaswamy.
\newblock Cognitive radar framework for target detection and tracking.
\newblock {\em IEEE Journal of Selected Topics in Signal Processing},
  9(8):1427--1439, 2015.

\bibitem{ECCMsurvey}
L.~Neng-Jing and Z.~Yi-Ting.
\newblock A survey of radar ecm and eccm.
\newblock {\em IEEE Transactions on Aerospace and Electronic Systems},
  31(3):1110--1120, 1995.

\bibitem{ECCM1}
C.~Shi, F.~Wang, M.~Sellathurai, and J.~Zhou.
\newblock Low probability of intercept-based distributed mimo radar waveform
  design against barrage jamming in signal-dependent clutter and coloured
  noise.
\newblock {\em IET Signal Processing}, 13(4):415--423, 2019.

\bibitem{ECCM2}
F.~A. Butt, I.~H. Naqvi, and U.~Riaz.
\newblock Hybrid phased-mimo radar: A novel approach with optimal performance
  under electronic countermeasures.
\newblock {\em IEEE Communications Letters}, 22(6):1184--1187, 2018.

\bibitem{stealth1}
W.-Q. Wang.
\newblock Moving-target tracking by cognitive rf stealth radar using frequency
  diverse array antenna.
\newblock {\em IEEE Transactions on Geoscience and Remote Sensing},
  54(7):3764--3773, 2016.

\bibitem{stealth2}
W.-Q. Wang.
\newblock Adaptive rf stealth beamforming for frequency diverse array radar.
\newblock In {\em 2015 23rd European Signal Processing Conference (EUSIPCO)},
  pages 1158--1161. IEEE, 2015.

\bibitem{stealth3}
Z.~Zhang, S.~Salous, H.~Li, and Y.~Tian.
\newblock Optimal coordination method of opportunistic array radars for
  multi-target-tracking-based radio frequency stealth in clutter.
\newblock {\em Radio Science}, 50(11):1187--1196, 2015.

\bibitem{Afr87}
S.~Afriat.
\newblock {\em Logic of choice and economic theory}.
\newblock Clarendon Press Oxford, 1987.

\bibitem{FM09}
F.~Forges and E.~Minelli.
\newblock Afriat's theorem for general budget sets.
\newblock {\em Journal of Economic Theory}, 144(1):135--145, 2009.

\bibitem{Var12}
H.~Varian.
\newblock Revealed preference and its applications.
\newblock {\em The Economic Journal}, 122(560):332--338, 2012.

\bibitem{Var06}
H.~Varian.
\newblock Revealed preference.
\newblock {\em Samuelsonian economics and the twenty-first century}, pages
  99--115, 2006.

\bibitem{Var82}
H.~Varian.
\newblock The nonparametric approach to demand analysis.
\newblock {\em Econometrica}, 50(1):945--973, 1982.

\bibitem{PK21unifying}
K.~Pattanayak and V.~Krishnamurthy.
\newblock Unifying classical and bayesian revealed preference.
\newblock {\em arXiv preprint arXiv:2106.14486}, 2021.

\bibitem{KPGKR21}
V.~Krishnamurthy, K.~Pattanayak, S.~Gogineni, B.~Kang, and M.~Rangaswamy.
\newblock Adversarial radar inference: Inverse tracking, identifying cognition,
  and designing smart interference.
\newblock {\em IEEE Transactions on Aerospace and Electronic Systems},
  57(4):2067--2081, 2021.

\bibitem{FST04}
A.~Fostel, H.~Scarf, and M.~Todd.
\newblock Two new proofs of {Afriat’s} theorem.
\newblock {\em Economic Theory}, 24(1):211--219, 2004.

\end{thebibliography}

\end{document}